\documentclass[runningheads]{llncs}
\usepackage[T1]{fontenc}
\usepackage{graphicx}
\usepackage{amsmath}
\usepackage{amssymb}
\usepackage{booktabs}
\usepackage{multirow}
\usepackage{subcaption}
\usepackage[most]{tcolorbox}
\usepackage{hyperref}
\usepackage{color}

\usepackage{latexsym}
\usepackage{bbm}
\usepackage{enumitem}
\usepackage{algorithm}
\usepackage{algorithmic}

\usepackage[utf8]{inputenc}

\usepackage{microtype}

\begin{document}
\title{What properties of reasoning supervision are associated with improved downstream model quality?}
\titlerunning{Reasoning Supervision Properties and Model Quality}

\author{Mikołaj Langner\orcidID{0009-0007-9531-5329} \and 
Dzmitry Pihulski\orcidID{0009-0002-5434-4696} \and
Jan Eliasz\orcidID{0009-0007-0851-1816} \and
Michał Rajkowski\orcidID{0009-0003-2983-4324} \and
Przemysław Kazienko\orcidID{0000-0001-5868-356X} \and
Maciej Piasecki\orcidID{0000-0003-1503-0993} \and
Jan Kocoń\orcidID{0000-0002-7665-6896} \and
Teddy Ferdinan\orcidID{0000-0003-3701-3502}}

\authorrunning{M. Langner et al.}
\institute{Wroclaw Tech, 50-370 Wrocław, Poland\\
\email{\{mikolaj.langner, dzmitry.pihulski, jan.eliasz, michal.rajkowski, kazienko, maciej.piasecki, jan.kocon, teddy.ferdinan\}@pwr.edu.pl}}
\maketitle
\begin{abstract}
Validating training data for reasoning models typically requires expensive trial-and-error fine-tuning cycles. In this work, we investigate whether the utility of a reasoning dataset can be reliably predicted prior to training using intrinsic data metrics. We propose a suite of quantitative measures and evaluate their predictive power by fine-tuning 8B and 11B models on semantically distinct variants of a Polish reasoning dataset. Our analysis reveals that these intrinsic metrics demonstrate strong and significant correlations with downstream model performance. Crucially, we find that the predictors of utility are scale-dependent: smaller models rely on alignment-focused metrics to ensure precision, whereas larger models benefit from high redundancy, utilizing verbose traces to solve complex tasks. These findings establish a scale-aware framework for validating reasoning data, enabling practitioners to select effective training sets without the need for exhaustive empirical testing.

\end{abstract}

\section{Introduction}
\label{sec:intro}

Explicit reasoning strategies \cite{wei2022chain} and specialized models \cite{openaio3,guo2025deepseek} have transformed the capabilities of Large Language Models (LLMs). Consequently, fine-tuning on datasets enriched with reasoning traces has become the standard paradigm for imbuing these models with such skills. However, while the importance of high-quality data is universally acknowledged, the definition of quality for reasoning traces remains ambiguous.

Currently, validating a reasoning dataset is an inefficient process that relies on post-hoc evaluation: researchers must fine-tune a model to discover if their data is effective. This \textit{training-as-validation} approach is computationally prohibitive and unscalable. To democratize the development of robust reasoning models, the community requires objective and computable metrics that can validate the utility of training data \textbf{before} the expensive fine-tuning process begins.

In this paper, we address this gap by establishing a link between intrinsic data characteristics and downstream model performance. We leverage a controlled set of Polish reasoning variants from our previous work~\cite{pihulski-etal-2026-breaking} and the corresponding fine-tuned 8B and 11B models. By subjecting these known reasoning variants to a rigorous set of quantitative measurements, from linguistic complexity to semantic alignment, we determine which metrics serve as reliable predictors of a model's final reasoning ability.

Our analysis is guided by the following research questions:

\begin{enumerate}[label={{\textbf{RQ\arabic*}}:},leftmargin=1.5cm, nosep]
    \item Is it feasible to validate the utility prior to fine-tuning?
    \item Which specific quantitative measures provide the most meaningful signal for validating training data quality?
\end{enumerate}

The contributions of this work are as follows: (1) a systematic evaluation of validation metrics for reasoning datasets, distinguishing between superficial statistics and deep semantic indicators; (2) a correlation analysis linking pre-training data scores with downstream performance; and, (3) a scale-aware framework for selecting reasoning data, allowing researchers to estimate model performance without incurring the full cost of training.

\subsection{Related Work}

The precise utility of generated \textit{reasoning traces} remains a subject of active debate. Shojaee et al.~\cite{shojaee2025_illusion} argue that reasoning-augmented models often exhibit \textit{illusionary} improvements, failing catastrophically on complex tasks while \textit{overthinking} simple ones. Although Lawsen et al.~\cite{lawsen2025_commentonillusion} challenged these findings based on methodological discrepancies, the consensus remains that reasoning traces are not a guaranteed panacea. Furthermore, studies indicate that LLM reasoning often diverges from genuine logical inference~\cite{chua2025deepseek,chen2025reasoning,kambhampati2025stop,wozniak2024personalized,chang2025global,ferdinan2025architectural,chodak2025typology,pihulski-etal-2026-breaking}, with models frequently omitting premises or generating hallucinated reasoning steps that do not correlate with the accuracy of the final answer.

Recent work has attempted to isolate specific attributes of reasoning data that drive performance, particularly sequence length. Jin et al.~\cite{jin2024-length} posit that extending the length of reasoning, regardless of quality, can boost performance. In contrast, Wu et al.~\cite{wu2025more} demonstrate an inverted U-shaped relationship, suggesting that excessive length introduces error accumulation. \looseness=-1

Collectively, these conflicting findings suggest that neither \textit{length} nor \textit{presence of reasoning} alone are sufficient proxies for the utility of training data. Although prior work largely evaluates reasoning quality by analyzing model outputs, there is insufficient research on validation methods that evaluate reasoning data before committing computational resources to fine-tuning. Our work addresses this gap by correlating intrinsic data metrics with downstream performance established in our prior experiments.

\begin{figure*}[htbp]
  \centering
  \fbox{\includegraphics[width=0.98\textwidth]{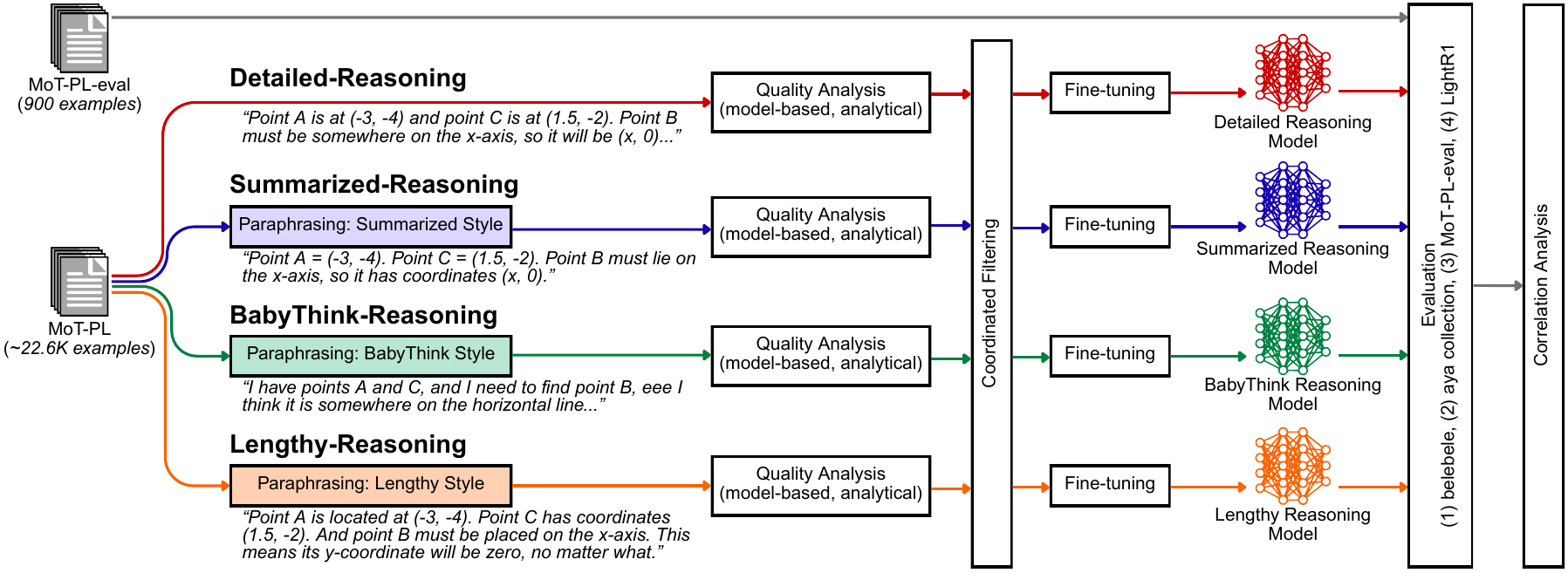}}
  \caption{We translated a subset of Mixture-of-Thoughts~\cite{mot2025_openr1,mot2025_math,mot2025_code,mot2025_science} into Polish, and split it into a training (MoT-PL) and evaluation set (MoT-PL-eval). Three additional variants of MoT-PL were created by paraphrasing only the reasoning part of each example: the \textit{Summarized} style made the reasoning much more concise, the \textit{BabyThink} style greatly simplified the reasoning, and the \textit{Lengthy} style prolonged the reasoning. Afterwards, we fine-tuned PLLuM-8B-instruct and Bielik-11B-v2.6-Instruct on these datasets separately and evaluated them.}
  \label{fig:illustration}
\end{figure*}

\section{Experimental Setup}
\label{sec:experimental_setup}

\subsection{Datasets}
\label{sec:datasets}

To rigorously evaluate the efficacy of pre-training validation metrics, four distinct reasoning datasets derived from the \textbf{Polish Mixture-of-Thoughts (MoT-PL)} were used. 
The original MoT-PL dataset was created by sampling approximately 32,000 examples from the English Mixture-of-Thoughts collection \cite{mot2025_openr1} and translating them into Polish using DeepSeek-V3 \cite{deepseekai2024}. After filtering for errors and context length, the final dataset contained 22,571 examples spanning three domains: Mathematics (28\%), Programming (17\%), and Science (55\%). To ensure the generated traces exhibited natural, human-like fluency rather than rigid machine translation artifacts, a randomly sampled subset of the DeepSeek-V3 outputs was manually verified by native Polish speakers.

From this foundational dataset, we generated four semantically distinct variants to serve as our controlled variables (see Figure~\ref{fig:illustration}). These datasets, \textit{Detailed}, \textit{Summarized}, \textit{BabyThink}, and \textit{Lengthy}, share identical user prompts and final answers but differ significantly in the style, length, and semantic density of their reasoning traces. The general statistics of the dataset variants are shown in Table~\ref{tab:dataset_stats}. The variants were generated by automatic paraphrasing using DeepSeek-V3, resulting in the following profiles:

\begin{itemize}[nosep, leftmargin=*]
    \item \textbf{Detailed:} The unmodified MoT-PL dataset, representing high-quality standard reasoning. The traces mimic the depth of the original English Mixture-of-Thoughts, serving as our control for \textit{standard} reasoning density.
    
    \item \textbf{Summarized:} A concise variant in which reasoning traces were compressed to retain essential logic while stripping stylistic fluff. This dataset tests the hypothesis that a higher information density correlates with efficiency.\looseness=-1
    
    \item \textbf{BabyThink:} A variant paraphrased into "childlike" language. Rather than merely reducing statistical readability, the prompt intentionally obfuscates specific details and calculations with vague filler. The original train of thought and structure are strictly preserved to avoid injecting artificial hallucinations or new reasoning fallacies.
    
    \item \textbf{Lengthy:} An artificially prolonged variant designed to be approximately twice as long as the \textit{Detailed} version. It preserves the original logic but introduces verbosity, allowing us to test if metrics favoring longer chains are misleading.
\end{itemize}

\begin{table}[htbp]
  \centering
  \caption{Statistical profile of the dataset variants used for metric validation. All variants share identical question/answer pairs; variations occur strictly within the reasoning trace. The first value of a token count comes from using the PLLuM-8B-instruct tokenizer, while the second value comes from using the Bielik-11B-v2.6-Instruct tokenizer.}
  \label{tab:dataset_stats}
  \small
  \begin{tabular}{lcccc}
    \toprule
    \textbf{Metric (Avg.)} & \textbf{Detailed} & \textbf{Summarized} & \textbf{BabyThink} & \textbf{Lengthy} \\
    \midrule
    Reasoning Tokens \hspace*{1em} & 2,613\textasciitilde 2,918 & 399\textasciitilde 452 & 3,729\textasciitilde 4,000 & 9,930\textasciitilde 10,902 \\
    Reasoning Chars \hspace*{1em} & 6,334 & 965 & 8,991 & 17,369 \\
    \midrule
    Total Seq. Tokens \hspace*{1em} & 3,288\textasciitilde 3,685 & 1,073\textasciitilde 1,219 & 4,404\textasciitilde 4,767 & 10,604\textasciitilde 11,669 \\
    Total Seq. Chars \hspace*{1em} & 8,090 & 2,721 & 10,747 & 25,938 \\
    \bottomrule
    \end{tabular}
\end{table}

All examples exceeding the context window limit (32k tokens) were filtered out prior to statistical analysis and training to ensure a consistent evaluation across all variants.

\subsection{Target Models}
To establish a robust performance baseline across different architectures, we utilized two state-of-the-art Polish-centric LLMs as a backbone for our experiments:\looseness=-1

\begin{itemize}[nosep, leftmargin=*]
    \item \textbf{PLLuM-8B-instruct} \cite{pllum2025,pkezik2025pllum,langner2025dichotomic}: A derivative of Llama-3.1-8B \cite{llama3models}, adapted via continual pre-training and instruction tuning on a massive Polish corpus;
    \item \textbf{Bielik-11B-v2.6-Instruct} \cite{ociepa2025bielik11bv2technical}: Built upon Mistral 7B v0.2 \cite{jiang2023mistral7b}, similarly enhanced with Polish-specific pre-training and fine-tuning.
\end{itemize}

Since none of the model possesses native reasoning capabilities, we adapted them by introducing special tokens \texttt{<think>} and \texttt{</think>} and expanding their embedding layers accordingly. The models were fine-tuned separately on four dataset variants (Section \ref{sec:datasets}), resulting in a diverse set of checkpoints with varying reasoning behaviors. The technical specifications are detailed in Appendix \ref{app:training_hparams}.

\subsection{Downstream Performance Benchmarks}
To measure the utility of the training data variants, we evaluated the fine-tuned models on a comprehensive suite of benchmarks. These evaluation scores serve as \textbf{ground truth labels} against which we correlate our pre-training data metrics.\looseness=-1

We selected four diverse benchmarks to capture different aspects of reasoning and language understanding:
\begin{itemize}[nosep, leftmargin=*]
    \item \textbf{MoT-PL-eval}: The held-out test split of our \textbf{MoT-PL} dataset (see Section~\ref{sec:datasets}), serving as the primary metric in-domain for Polish reasoning.
    \item \textbf{Belebele} \cite{2024-belebele}: A challenging multilingual reading comprehension benchmark testing the models' ability to extract information from complex passages.
    \item \textbf{Aya Collection} \cite{2024aya}: A broad instruction-following suite covering summarization, classification, and QA, used to verify general capability retention.
    \item \textbf{LightR1} \cite{lightR1}: An English-language benchmark for high-difficulty logical tasks, included to assess the transfer of cross-lingual reasoning.
\end{itemize}

\subsection{Evaluation Protocol}
\label{sec:eval_protocol}

To obtain the ground-truth performance scores needed for our correlation analysis, we evaluated all fine-tuned models on the four benchmarks described above. For each dataset, we sampled a stratified test set of 900 examples to ensure balanced coverage of reasoning lengths and task types. We report each model performance using two primary metrics: Absolute Accuracy and Relative Percentage Change compared to the base model, to isolate the specific impact of training data.\looseness=-1

Given the scale of evaluation, we adopted the \textit{LLM-as-a-judge} paradigm. We used \texttt{DeepSeek-R1-0528}~\cite{guo2025deepseek} as an oracle judge. The judge was strictly prompted to assess the correctness of the final answer (ignoring intermediate reasoning steps) against the ground truth. This binary decision process was applied across all benchmarks.

To ensure the reliability of these generated scores, we conducted a manual audit on a subset of 100 random samples from the \textbf{MoT-PL-eval} dataset. A human expert annotated these samples blindly (without seeing the model's judgment). The agreement rate between the human annotator and \texttt{DeepSeek-R1-0528} was \textbf{95\%}, with a Cohen’s Kappa score of \textbf{0.886}. This strong alignment confirms that our automated ground-truth labels are a reliable proxy for human evaluation.

During evaluation, the judge was provided with the query, reference answer, and model prediction, and instructed to output a binary decision in a constrained JSON format. The exact prompt templates for all benchmarks are available in our public repository\footnote{\url{https://github.com/DzmitryPihulski/prompts}}.

\section{Methodology}
\label{sec:methodology}

To systematically evaluate the utility of reasoning data prior to training, we propose a multi-dimensional validation framework. We categorize our metrics into two distinct groups: \textbf{Model-based Metrics} and \textbf{Analytical Metrics}. With Model-based Metrics, we aim to assess the logical integrity of the reasoning trace. We adopted the FVCU (Factuality, Validity, Coherence, Utility) taxonomy proposed by~\cite{lee-hockenmaier-2025-evaluating} for these metrics. Meanwhile, we designed our Analytical Metrics to measure statistical and structural properties of the text.\looseness=-1

\subsection{Model-based Metrics}
\label{sec:fvcu_method}
To assess the intrinsic quality of the reasoning steps beyond binary correctness, we implement an automated evaluation pipeline based on the \textbf{FVCU} taxonomy (Factuality, Validity, Coherence, Utility)~\cite{lee-hockenmaier-2025-evaluating}. This approach verifies whether the reasoning process itself is sound at the atomic level.

We utilize a two-stage pipeline consisting of an \textit{Atomizer} and a \textit{Judge}, both powered by \textbf{Qwen3-235B-A22B-Instruct-2507-FP8} \cite{qwen3technicalreport}.
\begin{enumerate}[nosep]
    \item \textbf{Atomizer:} Decomposes raw reasoning traces into atomic steps using a strict verbatim extraction strategy. This preserves the original density and style of the text, aligning with process supervision standards \cite{lightman2023letsverifystepstep}.
    \item \textbf{Judge:} Evaluates each step, one-by-one, against the FVCU taxonomy.
\end{enumerate}

\paragraph{Metric Definitions}
\begin{itemize}[nosep, leftmargin=*]
    \item \textbf{Factuality ($F$):} Assesses the consistency with premises and external truths using the \textit{Principal Knowledge Grounding} method \cite{hwang-etal-2025-assessing}, ensuring that steps are supported by explicit problem statements rather than hallucinated constraints.\looseness=-1
    \item \textbf{Validity ($V$):} Evaluates the mathematical and inferential correctness of the derivation. It distinguishes between calculation errors and logical fallacies.
    \item \textbf{Coherence ($C$):} Checks if the step logically follows the preceding one without gaps, satisfying the Markov property of the chain \cite{teng2025atomthoughtsmarkovllm}.
    \item \textbf{Utility ($U$):} Measures whether the step contributes effective progress towards the solution, distinguishing constructive decomposition from "reasoning loops".\looseness=-1
\end{itemize}

\subsection{Analytical Metrics}
\label{sec:auto_metrics}

To complement the computationally expensive FVCU, we compute scalable structural metrics across the full training dataset:

\begin{itemize}[nosep, leftmargin=*]
    \item \textbf{Semantic Alignment:} Cosine similarity between query and reasoning trace embeddings (using \texttt{mmlw-roberta-large}~\cite{dadas2024pirb}), serving as a proxy for instruction adherence~\cite{zanotto2025_linguistic}.
    \item \textbf{Semantic Flow:} Average cosine similarity between consecutive sentences, quantifying narrative smoothness, and transitional logic~\cite{zanotto2025_linguistic}.
    \item \textbf{Redundancy Ratio:} Information density calculated as $\left(1 - \frac{\text{len}_{compressed}}{\text{len}_{original}}\right)$, using \texttt{zlib} compression. Higher values indicate repetitive patterns or verbosity~\cite{rae2021_gopher,chen2024_datajuicer}.\looseness=-1
    \item \textbf{Syntactic Depth:} Average maximum depth of dependency trees (computed via \texttt{spacy} library), indicating linguistic complexity and cognitive load~\cite{zanotto2025_linguistic}.
    \item \textbf{Symbolic Fraction:} Ratio of non-alphanumeric characters to total text, capturing the density of mathematical or code-like notation~\cite{rae2021_gopher,chen2024_datajuicer}.
    \item \textbf{Perplexity:} Exponentiated average negative log-likelihood per token taken from \texttt{Qwen3-4B}~\cite{qwen3technicalreport}, measuring the text's conformity to general knowledge~\cite{rae2021_gopher,chen2024_datajuicer}.\looseness=-1
\end{itemize}

A core motivation of our framework is replacing costly trial-and-error fine-tuning with efficient pre-training validation. While brute-force empirical validation requires heavy forward and backward passes across all candidate datasets, incurring massive computational debt, our analytical pipeline bypasses gradient updates entirely. By relying strictly on lightweight processing and single-pass embedding extraction, we reduce the validation footprint from dozens of multi-GPU hours to negligible compute time.

\section{Results and Analysis}
\label{sec:results}

In this section, we present the empirical findings of our validation study. We begin by analyzing the intrinsic quality of the datasets using our framework: first, the model-based evaluation on a 1,000 subsample and second, the analytical profiling of the full training corpora. Finally, we report the performance of the model in downstream tasks and correlate these metrics to identify the most reliable predictors of success.\looseness=-1

\subsection{Model-based Metrics}
\label{sec:fvcu_results}

Due to the prohibitive computational cost of model-based judging, FVCU metrics were evaluated on a single subsample of 1,000 examples per variant. To mitigate the variance inherent in single-batch evaluation, we employed rigorous stratified sampling, ensuring the subset accurately preserves the domain and complexity distribution of the full dataset. While the lack of multiple independent batches precludes formal variance calculations, this stratified design yields a highly representative estimate. Consequently, we frame these FVCU scores not as absolute statistical bounds, but as robust directional indicators of the reasoning trade-offs between our dataset variants. Table~\ref{tab:fvcu_results} presents these results.

\begin{table}[h]
\centering
\caption{Model-based metrics evaluation on 1,000 \textbf{MoT-PL} subsamples.}
\label{tab:fvcu_results}
\renewcommand{\arraystretch}{0.8}
\small
\setlength{\tabcolsep}{6pt}
\begin{tabular}{l c c c c}
\toprule
\textbf{Dataset} & \textbf{Factuality} & \textbf{Validity} & \textbf{Coherence} & \textbf{Utility} \\
\midrule
\textbf{BabyThink} & 78.0 & 65.8 & 91.4 & 54.4 \\
\textbf{Detailed} & 95.0 & 94.3 & 97.5 & 84.5 \\
\textbf{Lengthy} & 94.8 & 95.2 & 98.5 & 87.0 \\
\textbf{Summarized} & 92.2 & 87.3 & 96.7 & 90.5 \\
\bottomrule
\end{tabular}
\end{table}

The \textbf{Summarized} variant maximizes Utility (90.5\%) but at the expense of Validity (87.3\%). This expected drop in Validity occurs because our LLM judge strictly evaluates explicit step-by-step derivation, penalizing the intentional omission of intermediate steps as logical gaps even when the final conclusion remains factual and highly useful. In contrast, \textbf{Lengthy} achieves the highest Validity (95.2\%) and Coherence (98.5\%), indicating that granular, explicit derivations are essential for stabilizing the reasoning process. Finally, the baseline \textbf{BabyThink} demonstrates that high Coherence (91.4\%) is insufficient for reasoning quality --- its low Validity (65.8\%) confirms that the model can generate linguistically smooth but factually ungrounded chains.\looseness=-1

These findings highlight a critical trade-off in reasoning data curation: while stripping intermediate steps (\textbf{Summarized}) increases immediate task utility, it degrades the rigorous logical grounding required for out-of-distribution generalization. Conversely, verbosity (\textbf{Lengthy}) acts as a safeguard against hallucination by enforcing strict state-tracking, which is essential for complex reasoning but requires sufficient model capacity to process.

\subsection{Analytical Metrics}
\label{sec:auto_results}

We extended our analysis to the entire training dataset using computationally efficient metrics. Table~\ref{tab:auto_metrics} summarizes the profiles of each variant.

\begin{table*}[htbp]
  \centering
  \caption{Analytical metrics calculated on the full training \textbf{MoT-PL} datasets.}
  \label{tab:auto_metrics}
  \renewcommand{\arraystretch}{0.8}
  \small
  \begin{tabular}{lcccccc}
    \toprule
    \textbf{Dataset} & \textbf{Syntactic} & \textbf{Sem.} & \textbf{Sem.} & \textbf{Perplexity} & \textbf{Redundancy} & \textbf{Symbolic} \\
     & \textbf{Depth} & \textbf{Flow} & \textbf{Align.} & & \textbf{Ratio} & \textbf{Fraction} \\
    \midrule
    \textbf{BabyThink} & 3.22 & 0.861 & 0.916 & 2.42 & 0.622 & 0.098 \\
    \textbf{Detailed} & 4.41 & 0.868 & 0.951 & 1.41 & 0.623 & 0.131 \\
    \textbf{Lengthy} & 4.61 & 0.856 & 0.941 & 1.76 & 0.629 & 0.127 \\
    \textbf{Summarized} & 4.92 & 0.881 & 0.950 & 1.38 & 0.441 & 0.201 \\
    \bottomrule
  \end{tabular}
\end{table*}

The \textit{Summarized} dataset emerges as the most information-dense, exhibiting the highest \textbf{Symbolic Fraction} (0.201) and \textbf{Syntactic Depth} (4.92) while maintaining the lowest \textbf{Redundancy Ratio} (0.441). In contrast, the \textit{Lengthy} and \textit{Detailed} variants share nearly identical redundancy scores ($\sim$0.62), suggesting that the \textit{Lengthy} variant scales volume without altering the fundamental compression rate of the text. Notably, \textit{BabyThink} variant, despite its simplified vocabulary, yields the highest \textbf{Perplexity} (2.42) and the lowest \textbf{Semantic Alignment} (0.916).\looseness=-1

These structural differences imply that reasoning quality is not merely a function of length but of information pacing. The high Perplexity and low Semantic Alignment of the \textit{BabyThink} variant suggest that artificially simplifying vocabulary disrupts the natural language distribution the model expects, paradoxically making the reasoning harder to learn from despite its simpler syntax.

\subsection{Downstream Model Performance}
\label{sec:model_results}

We evaluate the fine-tuned models across four benchmarks to establish the ground truth for our correlation analysis. We present the results in three stages: Absolute Accuracy, Relative Performance Change, and finally domain-specific breakdown.

Table \ref{tab:absolute_results} presents the absolute accuracy. Consistent with the difference in model size, Bielik-11B significantly outperforms PLLuM-8B. For \textbf{PLLuM-8B}, the \textit{Detailed} variant achieves the highest average performance (0.513), showing particular strength in Polish reasoning tasks on MoT-PL-eval (0.374). For \textbf{Bielik-11B}, the \textit{Lengthy} variant emerges as the superior specialist in reasoning overall, achieving the highest absolute scores on both MoT-PL-eval (0.701) and LightR1 (0.599).\looseness=-1

\begin{table*}[h]
    \centering
    \caption{Absolute Accuracy on downstream tasks. \textbf{Avg.} is the macro-average.}
    \label{tab:absolute_results}
    \small
    \setlength{\tabcolsep}{2.5pt}
    \renewcommand{\arraystretch}{0.8}
    \begin{tabular}{lcccccc|c}
        \toprule
        \textbf{Model Variant} & \textbf{Bel-PL} & \textbf{Bel-EN} & \textbf{Aya-PL} & \textbf{Aya-EN} & \textbf{MoT-PL} & \textbf{LightR1} & \textbf{Avg.} \\
        \midrule
        \multicolumn{8}{l}{\textit{PLLuM-8B-Instruct}} \\
        \quad Original & 0.609 & 0.656 & \textbf{0.656} & 0.552 & 0.316 & \textbf{0.172} & 0.494 \\
        \quad Detailed & \textbf{0.672} & \textbf{0.742} & 0.615 & \textbf{0.583} & \textbf{0.374} & 0.094 & \textbf{0.513} \\
        \quad Summarized & 0.623 & 0.673 & 0.572 & 0.486 & 0.352 & 0.070 & 0.463 \\
        \quad BabyThink & 0.550 & 0.628 & 0.487 & 0.389 & 0.305 & 0.071 & 0.405 \\
        \quad Lengthy & 0.512 & 0.557 & 0.364 & 0.279 & 0.309 & 0.070 & 0.349 \\
        \midrule
        \multicolumn{8}{l}{\textit{Bielik-11B-v2.6-Instruct}} \\
        \quad Original & 0.876 & 0.904 & 0.856 & \textbf{0.903} & 0.624 & 0.521 & 0.781 \\
        \quad Detailed & \textbf{0.894} & \textbf{0.937} & \textbf{0.861} & 0.854 & 0.671 & 0.537 & \textbf{0.792} \\
        \quad Summarized & 0.826 & 0.876 & 0.826 & 0.872 & 0.529 & 0.264 & 0.699 \\
        \quad BabyThink & 0.810 & 0.871 & 0.757 & 0.867 & 0.497 & 0.266 & 0.678 \\
        \quad Lengthy & 0.819 & 0.841 & 0.772 & 0.891 & \textbf{0.701} & \textbf{0.599} & 0.771 \\
        \bottomrule
    \end{tabular}
\end{table*}

Table \ref{tab:relative_change} reports the Relative Percentage Change to normalize for the base model capabilities. The \textbf{Detailed} model proved to be the safest strategy, delivering consistent gains for PLLuM-8B on most benchmarks. Including general NLP benchmarks in the average demonstrates that high-quality reasoning (\textbf{Detailed}) improves standard tasks. However, isolated reasoning benchmarks reveal an expected trade-off: fine-tuning exclusively on MoT-PL boosts our target Polish reasoning but degrades English reasoning (LightR1) due to mild catastrophic forgetting of English chain-of-thought capabilities. Finally, the \textbf{Lengthy} dataset exhibits a volatile profile: on the smaller PLLuM-8B, it caused catastrophic forgetting, but in the larger Bielik-11B model, it unlocked significant reasoning capabilities, increasing performance on MoT-PL-eval by +12.3\% and LightR1 by +15.0\%.\looseness=-1

\begin{table*}[h]
    \centering
    \caption{Relative Percentage Change (\%) on downstream tasks between original and finetuned models.}
    \label{tab:relative_change}
    \small
    \setlength{\tabcolsep}{2.5pt}
    \renewcommand{\arraystretch}{0.8}
    \begin{tabular}{lcccccc|c}
        \toprule
        \textbf{Model} & \textbf{Bel-PL} & \textbf{Bel-EN} & \textbf{Aya-PL} & \textbf{Aya-EN} & \textbf{MoT-PL} & \textbf{LightR1} & \textbf{Avg.} \\
        \midrule
        \multicolumn{8}{l}{\textit{PLLuM-8B-Instruct}} \\
        \quad Detailed & \textbf{+10.3\%} & \textbf{+13.1\%} & -6.2\% & \textbf{+5.6\%} & \textbf{+18.4\%} & -45.3\% & \textbf{+3.8\%} \\
        \quad Summarized & +2.3\% & +2.6\% & -12.8\% & -12.0\% & +11.4\% & -59.3\% & -6.3\% \\
        \quad BabyThink & -9.7\% & -4.3\% & -25.8\% & -29.5\% & -3.5\% & -58.7\% & -18.0\% \\
        \quad Lengthy & -15.9\% & -15.1\% & -44.5\% & -49.5\% & -2.2\% & -59.3\% & -29.4\% \\
        \midrule
        \multicolumn{8}{l}{\textit{Bielik-11B-v2.6-Instruct}} \\
        \quad Detailed & \textbf{+2.1\%} & \textbf{+3.7\%} & \textbf{+0.6\%} & -5.4\% & +7.5\% & +3.1\% & \textbf{+1.4\%} \\
        \quad Summarized & -5.7\% & -3.1\% & -3.5\% & -3.4\% & -15.2\% & -49.3\% & -10.5\% \\
        \quad BabyThink & -7.5\% & -3.7\% & -11.6\% & -4.0\% & -20.4\% & -48.9\% & -13.2\% \\
        \quad Lengthy & -6.5\% & -7.0\% & -9.8\% & -1.3\% & \textbf{+12.3\%} & \textbf{+15.0\%} & -1.3\% \\
        \bottomrule
    \end{tabular}
\end{table*}

The domain-specific breakdown in Table \ref{tab:domain_merged} exposes a critical dependency between model capacity and reasoning density. In \textbf{MATH}, we observe a striking inversion of preferences: the smaller \textbf{PLLuM-8B} benefits exclusively from the \textit{Summarized} variant (+26.2\%), likely succumbing to context drift in longer chains, whereas the larger \textbf{Bielik-11B} effectively utilizes the "thinking space" of \textit{Lengthy} derivations (+12.5\%) to navigate complex logic. This capacity gap is most acute in \textbf{CODE}, where verbose reasoning acts as a crucial scaffold for Bielik-11B (+131.4\%) but induces catastrophic forgetting in PLLuM-8B (-73\% to -96\%). In contrast, in \textbf{SCIENCE}, the smaller model sees the largest relative gains (+28.6\%), suggesting that reasoning traces help unlock latent knowledge, while the larger model hits a performance ceiling with only marginal improvements (+5.0\%).\looseness=-1

\begin{table}[h]
    \centering
    \caption{MoT-PL-eval performance by domain. Cells show \textbf{Accuracy} followed by \textbf{(Relative Gain \%)} compared to the Original baseline. Bold indicates the best result per model/domain.}
    \label{tab:domain_merged}
    \small
    \setlength{\tabcolsep}{4pt}
    \renewcommand{\arraystretch}{0.8}
    \begin{tabular}{lccc}
        \toprule
        \textbf{Model} & \textbf{MATH} & \textbf{CODE} & \textbf{SCIENCE} \\
        \midrule
        \multicolumn{4}{l}{\textit{PLLuM-8B-Instruct}} \\
        \quad Original   & 0.103 & \textbf{0.138} & 0.479 \\
        \quad Detailed   & 0.100 (-2.9\%) & 0.020 (-85.5\%) & \textbf{0.616 (+28.6\%)} \\
        \quad Summarized & \textbf{0.130 (+26.2\%)} & 0.006 (-95.7\%) & 0.574 (+19.8\%) \\
        \quad BabyThink  & 0.058 (-43.7\%) & 0.012 (-91.3\%) & 0.523 (+9.2\%) \\
        \quad Lengthy    & 0.076 (-26.2\%) & 0.037 (-73.2\%) & 0.511 (+6.7\%) \\
        \midrule
        \multicolumn{4}{l}{\textit{Bielik-11B-v2.6-Instruct}} \\
        \quad Original   & 0.593 & 0.175 & 0.785 \\
        \quad Detailed   & 0.577 (-2.7\%) & 0.346 (+97.7\%) & \textbf{0.824 (+5.0\%)} \\
        \quad Summarized & 0.276 (-53.5\%) & 0.112 (-36.0\%) & 0.791 (+0.8\%) \\
        \quad BabyThink  & 0.285 (-51.9\%) & 0.127 (-27.4\%) & 0.722 (-8.0\%) \\
        \quad Lengthy    & \textbf{0.667 (+12.5\%)} & \textbf{0.405 (+131.4\%)} & 0.809 (+3.1\%) \\
        \bottomrule
    \end{tabular}
\end{table}

\subsection{Correlation Analysis: Drivers of Reasoning Performance}
\label{sec:correlation}

To understand the mechanisms behind the observed performance changes, we analyzed the relationship between our training data metrics (defined in Sections~\ref{sec:fvcu_method} and \ref{sec:auto_metrics}) and the downstream performance. We calculated the Spearman Rank Correlation ($\rho$) between each metric and the relative performance gain.

Table~\ref{tab:corr_global} highlights a distinct divergence in how training data characteristics translate to downstream performance. For \textbf{PLLuM-8B}, performance is primarily driven by \textbf{Semantic Alignment} ($\rho_{avg}=0.75$), \textbf{Semantic Flow} ($\rho_{avg}=0.65$) and \textbf{Factuality} ($\rho_{avg}=0.45$). This suggests that the smaller model relies heavily on clear, instruction-compliant data. In particular, \textbf{Utility} shows a strong negative correlation with the complex \textsc{LightR1} benchmark ($\rho=-0.74$). This indicates that data optimized for high utility, typically concise summaries, deprive the model of the intermediate reasoning tokens necessary to learn complex logic steps. In contrast, \textbf{Bielik-11B} demonstrates a strong dependence on reasoning volume and correctness. Although the \textbf{Redundancy Ratio} perfectly predicts the success in \textsc{LightR1} ($\rho=1.0$), the model-based metrics clarify the nature of this redundancy. \textbf{Validity} and \textbf{Coherence} show near-perfect correlations with reasoning tasks, confirming that the model leverages redundant tokens effectively only when they form a logically valid reasoning. In contrast, \textbf{Semantic Flow} correlates negatively with hard reasoning ($\rho=-0.80$), reinforcing that narrative smoothness is less critical than a rigorous step-by-step derivation for the larger model.\looseness=-1

\begin{table}[t]
\centering
\caption{Spearman's $\rho$ between training dataset metrics and downstream performance on general benchmarks. The metrics are divided into \textbf{Analytical} (on full dataset) and \textbf{Model-based} (on a stratified subsample of 1,000 examples).}
\label{tab:corr_global}
\small
\renewcommand{\arraystretch}{0.8}
\setlength{\tabcolsep}{2pt}
\begin{tabular}{lccccccc}
\toprule
\textbf{Metric} & \textbf{Bel-PL} & \textbf{Bel-EN} & \textbf{Aya-PL} & \textbf{Aya-EN} & \textbf{MoT-PL} & \textbf{LightR1} & \textbf{Avg.} \\
\midrule
\multicolumn{8}{l}{\textbf{\textit{PLLuM-8B-Instruct}}} \\
\midrule
\multicolumn{8}{l}{\textit{Analytical Metrics}} \\
Redundancy Ratio   & -0.40 & -0.40 & -0.40 & -0.40 & 0.00 & 0.11 & -0.25 \\
Semantic Alignment & \textbf{0.80} & \textbf{0.80} & \textbf{0.80} & \textbf{0.80} & \textbf{1.00} & 0.32 & \textbf{0.75} \\
Semantic Flow      & \textbf{0.80} & \textbf{0.80} & \textbf{0.80} & \textbf{0.80} & 0.60 & 0.11 & 0.65 \\
Symbolic Fraction  & 0.60 & 0.60 & 0.60 & 0.60 & 0.80 & -0.21 & 0.50 \\
Syntactic Depth    & 0.00 & 0.00 & 0.00 & 0.00 & 0.40 & \textbf{-0.74} & -0.06 \\
\midrule
\multicolumn{8}{l}{\textit{Model-based Metrics}} \\
Validity ($V$)     & -0.20 & -0.20 & -0.20 & -0.20 & 0.40 & -0.21 & -0.10 \\
Factuality ($F$)   & \textbf{0.40} & \textbf{0.40} & \textbf{0.40} & \textbf{0.40} & \textbf{0.80} & 0.32 & \textbf{0.45} \\
Coherence ($C$)    & -0.20 & -0.20 & -0.20 & -0.20 & 0.40 & -0.21 & -0.10 \\
Utility ($U$)      & 0.00 & 0.00 & 0.00 & 0.00 & 0.40 & \textbf{-0.74} & -0.06 \\
\midrule
\multicolumn{8}{l}{\textbf{\textit{Bielik-11B-v2.6-Instruct}}} \\
\midrule
\multicolumn{8}{l}{\textit{Analytical Metrics}} \\
Redundancy Ratio   & 0.00 & -0.40 & 0.00 & 0.20 & \textbf{0.80} & \textbf{1.00} & 0.27 \\
Semantic Alignment & \textbf{1.00} & \textbf{0.80} & \textbf{1.00} & -0.40 & 0.40 & 0.00 & \textbf{0.47} \\
Semantic Flow      & 0.60 & \textbf{0.80} & 0.60 & -0.40 & -0.40 & -0.80 & 0.07 \\
Symbolic Fraction  & 0.80 & 0.60 & 0.80 & 0.00 & 0.20 & -0.40 & 0.33 \\
Syntactic Depth    & 0.40 & 0.00 & 0.40 & \textbf{0.60} & 0.40 & -0.20 & 0.27 \\
\midrule
\multicolumn{8}{l}{\textit{Model-based Metrics}} \\
Validity ($V$)     & 0.40 & -0.20 & 0.40 & 0.40 & \textbf{1.00} & \textbf{0.80} & 0.47 \\
Factuality ($F$)   & \textbf{0.80} & \textbf{0.40} & \textbf{0.80} & -0.20 & 0.80 & 0.60 & \textbf{0.53} \\
Coherence ($C$)    & 0.40 & -0.20 & 0.40 & 0.40 & \textbf{1.00} & \textbf{0.80} & 0.47 \\
Utility ($U$)      & 0.40 & 0.00 & 0.40 & \textbf{0.60} & 0.40 & -0.20 & 0.27 \\
\bottomrule
\end{tabular}
\end{table}

Table~\ref{tab:corr_domain_horizontal} differentiates the drivers for procedural logic versus knowledge retrieval. In the \textsc{Code} and \textsc{Math} domains, \textbf{Semantic Flow} correlates negatively for Bielik-11B (reaching $\rho=-0.80$), indicating that narrative smoothness often impedes strict logical derivation.
For Bielik-11B, performance in these domains depends on a combination of reasoning volume and correctness. The model shows a perfect correlation with \textbf{Redundancy Ratio} ($\rho=1.0$) alongside strong correlations with \textbf{Validity} and \textbf{Coherence} ($\rho=0.80$). This suggests that the benefit of verbose reasoning comes from the generation of valid and coherent intermediate steps rather than redundancy alone.
PLLuM-8B shows a divergent pattern in \textsc{Math}, where performance correlates perfectly with \textbf{Symbolic Fraction} ($\rho=1.0$) and strongly with \textbf{Utility} ($\rho=0.80$), but weakly with \textbf{Validity} ($\rho=0.20$). This implies a reliance on formal notation and concise answers rather than the verification of the logical chain. However, in \textsc{Code}, PLLuM-8B aligns with the larger model, showing strong correlations with both \textbf{Redundancy} ($\rho=1.0$) and \textbf{Validity} ($\rho=0.80$).
Finally, \textsc{Science} is distinct; here, Bielik-11B exhibits a perfect correlation with \textbf{Factuality} ($\rho=1.0$), identifying factual accuracy as the sole critical driver, while PLLuM-8B relies primarily on \textbf{Semantic Flow} and \textbf{Semantic Alignment} ($\rho=0.80$).\looseness=-1

\begin{table}[t]
\centering
\caption{Spearman's $\rho$ on \textbf{Reasoning Domains} for \textbf{MoT-PL} dataset comparison. Left side: PLLuM-8B, Right side: Bielik-11B.}
\label{tab:corr_domain_horizontal}
\small
\renewcommand{\arraystretch}{0.8}
\setlength{\tabcolsep}{3pt}
\begin{tabular}{lcccccc}
\toprule
 & \multicolumn{3}{c}{\textbf{PLLuM-8B}} & \multicolumn{3}{c}{\textbf{Bielik-11B}} \\
\cmidrule(lr){2-4} \cmidrule(lr){5-7}
\textbf{Metric} & \textbf{MATH} & \textbf{CODE} & \textbf{SCIENCE} & \textbf{MATH} & \textbf{CODE} & \textbf{SCIENCE} \\
\midrule
\multicolumn{7}{l}{\textit{Analytical Metrics}} \\
Redundancy Ratio   & -0.40 & \textbf{1.00} & -0.40 & \textbf{1.00} & \textbf{1.00} & 0.60 \\
Semantic Alignment & 0.80 & 0.00 & \textbf{0.80} & 0.00 & 0.00 & \textbf{0.80} \\
Semantic Flow      & 0.80 & -0.80 & \textbf{0.80} & -0.80 & -0.80 & 0.00 \\
Symbolic Fraction  & \textbf{1.00} & -0.40 & 0.60 & -0.40 & -0.40 & 0.40 \\
Syntactic Depth    & 0.80 & -0.20 & 0.00 & -0.20 & -0.20 & 0.20 \\
\midrule
\multicolumn{7}{l}{\textit{Model-based Metrics}} \\
Validity ($V$)     & 0.20 & \textbf{0.80} & -0.20 & \textbf{0.80} & \textbf{0.80} & 0.80 \\
Factuality ($F$)   & 0.40 & 0.60 & \textbf{0.40} & 0.60 & 0.60 & \textbf{1.00} \\
Coherence ($C$)    & 0.20 & \textbf{0.80} & -0.20 & \textbf{0.80} & \textbf{0.80} & 0.80 \\
Utility ($U$)      & \textbf{0.80} & -0.20 & 0.00 & -0.20 & -0.20 & 0.20 \\
\bottomrule
\end{tabular}
\end{table}

\section{Discussion}
\label{sec:discussion}

\paragraph{RQ1. Is it feasible to validate the utility prior to fine-tuning?}
\textbf{Yes, but the predictive signal of the metrics depends the model size.}
Our analysis confirms that dataset metrics are reliable performance predictors ($\rho \ge 0.75$), yet there is no universal quality profile. For example, \textit{Redundancy Ratio} acts as a decisive positive signal for the Bielik-11B model in reasoning tasks ($\rho=1.0$) but remains neutral or negative for the PLLuM-8B model. Similarly, while \textit{Semantic Alignment} universally benefits general instruction following, it fails to predict success in complex reasoning for larger models. This indicates that pre-validation of training data requires a scale-based calibration; small models benefit more from semantic coherence with less redundancy, while larger models can more effectively leverage redundancy in longer reasoning trace.

\paragraph{RQ2. Which specific quantitative measures provide the most meaningful signal for validating training data quality?}
We observe a fundamental dichotomy in metric efficacy driven by the complexity threshold of the model.

For \textbf{PLLuM-8B}, performance is driven by \textit{Semantic Alignment} ($\rho=0.75$) and \textit{Factuality}. However, we observe a distinct negative correlation between \textit{Utility} and complex reasoning ($\rho=-0.74$ in LightR1). This suggests that data optimized for high human utility deprives smaller models of the intermediate tokens necessary to learn logic. Thus, for smaller models, the most critical signal is the directness and factual grounding of the data, rather than its reasoning depth.\looseness=-1

For \textbf{Bielik-11B}, \textit{Redundancy Ratio} is the strongest predictor of reasoning success ($\rho=1.0$), provided that it is supported by high \textit{Validity} ($\rho=0.80$). Crucially, \textit{Semantic Flow} correlates negatively with Math and Code performance ($\rho=-0.80$). This indicates that the larger model benefit from verbose, rigorous derivation steps, even if repetitive, rather than smooth narrative explanations. In knowledge-heavy domains like Science, this shifts entirely to \textit{Factuality} ($\rho=1.0$), rendering structural metrics less relevant.

\section{Conclusions}
\label{sec:conclusion}

This study establishes that effective data validation requires calibrating metrics to model capacity. By analyzing the correlation between the properties of the intrinsic data and the downstream performance, we identified distinct optimization requirements for different scales of parameters.

For a smaller model (PLLuM-8B), we observed a negative correlation between metrics favoring conciseness (Utility) and reasoning performance. These models rely primarily on \textit{Semantic Alignment} and \textit{Factuality} to prevent hallucinations, suggesting that training data should prioritize direct instruction adherence over complex reasoning chains.
In contrast, the larger model (Bielik-11B) demonstrated a strong positive correlation with \textit{Redundancy Ratio} in formal domains. This indicates that verbose iterative derivation steps are essential for performance on this scale. Consequently, data curation must distinguish between knowledge-intensive tasks, which benefit from factual density, and reasoning tasks, which require structural redundancy.

Building on these findings, future work will focus on extending this capacity-aware validation framework across a wider spectrum of model sizes to pinpoint the exact parameter threshold for reasoning verbosity. Additionally, we aim to employ instance-level influence functions to establish a direct causal link between specific structural data patterns and inference-time logical robustness. Simultaneously, we will investigate how much the reasoning setup affects other LLM properties, such as the tendency towards hallucination \cite{matys2025aggtruth} or in-context learning \cite{szczesny2025leveraging}.

\section{Limitations}
\label{sec:limitations}

Our findings suggest that 8B and 11B models use verbose reasoning data differently, but because we did not test intermediate or much larger models, we cannot tell whether this shift is gradual or appears at a specific scale. We also used disjoint train and test sets, which supports rigorous system-level correlation analysis but obscures instance-level effects, so we cannot identify how particular reasoning patterns affect individual predictions. In addition, our evaluation depends on an LLM judge, which may introduce bias despite strong agreement with human raters; reasoning variants generated with DeepSeek-V3 may contain paraphrasing artifacts that models can exploit; and testing Polish-fine-tuned models on English benchmarks introduces cross-lingual effects that make it harder to isolate reasoning ability from language processing limitations.

\begin{credits}
\subsubsection{\ackname}
This work was supported by: 
(1) the National Science Center, Poland, grant no. 2021/41/B/ST6/04471; 
(2) CLARIN-PL: Common Language Resources and Technology Infrastructure (POIR.04.02.00-00C002/19, 2024/WK/01, FENG.02.04-IP.040004/24); 
(3) Digital Research Infrastructure for the Arts and Humanities DARIAH-PL: POIR.04.02.00-00-D006/20, KPOD.01.18-IW.03-0013/23;
(4) the statutory funds of the Dept.of AI, Wroclaw Tech; 
(5) Polish Ministry of Education and Science: “International Projects Co-Funded”;
(6) the EU under the Horizon Europe, grant no. 101086321 (OMINO). The views expressed are those of the authors and do not necessarily reflect those of the EU or the European Research Executive Agency.

\subsubsection{\discintname}All authors 
% Mikołaj Langner, Dzmitry Pihulski, Jan Eliasz, Michał Rajkowski, Teddy Ferdinan, Jan Kocoń, Maciej Piasecki, and Przemysław Kazienko 
have received funding from the Ministry of Digital Affairs of Poland, Polish National Science Center, and the European Union.
\end{credits}

\bibliographystyle{splncs04}
\bibliography{main}

\section{Appendix}
\label{app:training_hparams}

Experiments were conducted on the WCSS LEM cluster\footnote{\url{https://www.wcss.pl/en/}} using nodes equipped with $4\times$ NVIDIA H100-94GB GPUs and Intel Xeon Platinum 8462Y+ CPUs. We utilized the \texttt{trl} library with DeepSpeed ZeRO Stage-3. Table~\ref{table_training_hparams} details the hyperparameters for both model families. We used the AdamW optimizer ($\beta_1=0.9, \beta_2=0.999, \epsilon=10^{-8}$). All model outputs were generated using fixed decoding strategies: temperature=0.6, top$-p$=0.95, top$-k$=20, min$-p$=0.1, and a repetition penalty of 1.2.

\begin{table}[h]
    \centering
    \caption{Fine-tuning hyperparameters used in our experiments.}
    \label{table_training_hparams}
    \small
    \setlength{\tabcolsep}{5pt}
    \renewcommand{\arraystretch}{0.8}
    \begin{tabular}{lcccccc}
        \toprule
        \textbf{Model} & \textbf{Epochs} & \textbf{Batch} & \textbf{Seq. Len} & \textbf{Peak LR} & \textbf{Sched.} & \textbf{Decay} \\
        \midrule
        PLLuM-8B & 2 & 128 & 8,192 & $4 \times 10^{-5}$ & Cosine & 0.1 \\
        Bielik-11B & 3 & 128 & 8,192 & $7 \times 10^{-6}$ & Linear & 0.0 \\
        \bottomrule
    \end{tabular}
\end{table}

\end{document}